\documentclass[11pt]{article}   
\usepackage{latex8}             
\usepackage{times}              
\usepackage{graphicx}           
\usepackage{url}
\usepackage{color}
\usepackage{amsfonts}
\begin{document}

\title{Automated Machine Learning  - a brief review at the end of the early years}
\author{Hugo Jair Escalante
\affiliation{Computer Science Department\\ Instituto Nacional de Astrof\'isica, \'Optica y Electr\'onica, Tonanzintla, Puebla, 72840, Mexico\\ \email{hugojair@inaoep.mx}}}
\maketitle

\begin{abstract}
Automated machine learning (AutoML) is the sub-field of machine learning that aims at automating, to some extend,  all stages of the design  of a machine learning system. In the context of supervised learning, AutoML is concerned with 
feature extraction, pre processing, model design and post processing. Major contributions and achievements in AutoML have been taking place during the recent decade. 
We are therefore in perfect timing to look back and realize what we have learned. This chapter aims to summarize the main findings in the early years of AutoML. More specifically, in this chapter  an introduction to AutoML for  supervised learning is provided and an historical review of progress in this field is presented. Likewise, the main paradigms of AutoML are  described and research opportunities are outlined. 
\end{abstract}


\section{Introduction}
\label{sec:intro} 
Automated Machine Learning or AutoML is a term coined by the machine learning community to refer to methods that aim at automating the design and development of machine learning systems and applications~\cite{DBLP:books/sp/HKV2019}.  In the context of supervised learning, AutoML aims at relaxing the need of the user in the loop from all stages in the design of supervised learning systems (i.e., any system relying on models for classification, recognition, regression, forecasting, etc.).
This is a tangible need at  present, as data are being generated vastly and in practically any context and scenario, however, the number of machine learning experts available to analyze such data is overseeded. 

AutoML for supervised learning has been the focus of research for more than ten years now\footnote{Please note that although model selection and other efforts for hyperparameter optimization have been out there for decades, see e.g.,~\cite{10.5555/1756006.1756009}; this chapter  focuses on full model or pipeline selection and design~\cite{psms06,DBLP:books/sp/HKV2019}.}, and great progress has been achieved so far, consider for instance the useful AutoML methods in the most popular machine learning toolkits~\cite{Feurer2019,ThoHutHooLey13-AutoWEKA}, and the AutoML mechanisms in large scale platforms (e.g., Azure~\footnote{\url{https://docs.microsoft.com/en-us/azure/machine-learning/service/concept-automated-ml}} or H2O.ai\footnote{\url{http://docs.h2o.ai/h2o/latest-stable/h2o-docs/automl.html}}~\cite{H2Opaper}). In fact, AutoML is nowadays a \emph{hot topic} within machine learning that is receiving much attention from industry, academy and even the general public. 

With such progress and interest from the community, it is necessary to go through the fundamentals and  main findings achieved in the last decade. This is the aim of the present chapter, which aims at 
reviewing the most notable developments in the last few years,  explaining the fundamentals of AutoML and highlighting open issues and research opportunities in the subject. 

This chapter is complimentary to excellent surveys and reviews in the field that can be found in~\cite{yao2018taking,zller2019survey,he2019automl,DBLP:books/sp/HKV2019,DBLP:journals/corr/abs-1907-08392,elsken2018neural,nelishia}. Compared with these references, this chapter offers an introduction to AutoML, and a brief review of progress in the field, all of this at a superficial but broad reaching focus. 

\textcolor{black}{The remainder of the chapter is organized as follows. In Section~\ref{sec:basics} the fundamentals of AutoML are introduced, including definitions, notions and components of AutoML systems. Then, in Section~\ref{sec:methods} a brief review on the most representative AutoML methodologies is presented. 
Next, in Section~\ref{sec:challenges}, a brief review on AutoML challenges and their role in the development of the fied are presented. 
Then in Section~\ref{sec:openissues} open issues and research opportunities are highlighted. Finally, in Section~\ref{sec:conclusions} a summary of the chapter and take-home messages are presented. }

\section{Fundamentals of AutoML}
\label{sec:basics}
Automated\footnote{Often referred too as Autonomous or Automatic Machine Learning}  Machine Learning (AutoML) is the field of study dealing with methods that aim at reducing the need of user interaction in the design of machine learning systems and applications. The topic has been mostly studied in the  context of supervised learning, although unsupervised~\cite{allingham2018a} and semi supervised learning~\cite{DBLP:conf/aaai/LiWWT19} efforts are emerging as well. This chapter deals with AutoML in the supervised learning context. 

\subsection{Supervised learning}
Supervised learning is perhaps the most studied topic within machine learning, as it has wide applicability. Spam filtering methods, face recognition systems, handwritten character recognition techniques and text classification methodologies are only a few of the \emph{classical } applications relying in supervised learning. The distinctive feature of supervised learning methods is that they must \emph{learn} to map objects to labels, based on a sample of labeled data (i.e., the supervision). 

More formally, under the supervised learning setting we have available a data set $\mathcal{D}$ formed by $N$ pairs of $d$-dimensional samples, $\mathbf{x}_i \in \mathbb{R}^d$, and labels\footnote{Please note that labels could be also real values (for regression tasks) or categorical, for clarity, we instead describe a binary classification problem.} $y_i \in \{-1, 1\}$, that is: $\mathcal{D} = \{ (\mathbf{x}_i, y_i) \}_{i \in 1, \ldots,N}$. The samples $\mathbf{x}_i$ codify objects of interest (e.g., documents, images or videos) with a set of numerical descriptors, while the labels $y_i$ determine the \emph{class} of objects (e.g., spam vs. no-spam). The overall goal of supervised learning is to find a function  $f: \mathbb{R}^d \rightarrow \{-1, 1\}$ mapping inputs to outputs, i.e., $y_j = f(\mathbf{x}_j)$, that can generalize beyond $\mathcal{D}$. Where options for the form of $f$ include linear models, decision trees, instance based classifiers among others. Regardless of the form of $f$, the learning process reduces to find the $f$ that  \emph{best} fits dataset $\mathcal{D}$.

Usually, $\mathcal{D}$ is split into training and validation partitions, hence the goal is learning $f$ from $\mathcal{D}$ such that label predictions can be made for any other instance sampled from the same underlying distribution as $\mathcal{D}$. If we denote  $\mathcal{T}$ to the \emph{test} set, formed by instances coming from the same distribution as $\mathcal{D}$ but that do not appear in such set. $\mathcal{T}$ can be used to evaluate the generalization capabilities of $f$. The reader is referred to~\cite{Alpaydin,hastie,bishop} for definitions and in depth treatments of supervised machine learning. 

\subsection{Notions of AutoML}
Having described the supervised learning setting, we can intuitively describe  AutoML as the \textbf{\emph{task of  finding the $f$ that better generalizes in any possible $\mathcal{T}$ with the less possible human intervention}}. Where $f$ can be the composition of multiple functions that may transform the input space, subsampling data, combining multiple predictors, etc. For example, $f$ could be of the form: $f(\textbf{x}) = \nu_{\theta_\nu}(\Phi_{\theta_\Phi}(\textbf{x})$, here $\nu$ is a classification model (e.g., a random forest classifier~\cite{randomforest}) and $\Phi$ is a feature transformation methodology (e.g., feature standardization and principal component analysis~\cite{guyonfs}) with hyperparameters $\theta_\nu$ and $\theta_\Phi$, respectively, and where each of these models could be formed in turn by several other functions/models.

Functions of the form $f(\textbf{x}) = \nu_{\theta_\nu}(\Phi_{\theta_\Phi}(\textbf{x})$ are called \emph{full models}~\cite{psms06,Escalante:2009:PSM:1577069.1577084} or \emph{pipelines}~\cite{Feurer2019}, as they comprise all of the processes that have to be applied to data in order to obtain a supervised learning model. 
AutoML can be seen as the search of functions $\nu$ and $\Phi$, with their corresponding hyperparameters  $\theta_\nu$ and $\theta_\Phi$ using $\mathcal{D}$. 
In the following we present conventional definitions of AutoML, however, the intuitive notion is general enough to be inclusive of all existing definitions, and it should be clearer for newcomers to the field. 

\subsubsection{Levels of automation in AutoML}
There are several notions of AutoML for supervised learning dating back to 2006 (see the \emph{Full model selection} definition\footnote{Although this article was published in 2009, the main concepts and ideas were presented in a NIPS workshop in 2006~\cite{psms06,DBLP:conf/ijcnn/GuyonSDC07}.} in~\cite{Escalante:2009:PSM:1577069.1577084}), where one of the mostly adopted  is that from Feurer et al.~\cite{Feurer2019}. Such definition, however, refers solely to the automatic pipeline generation problem, whereas different related tasks within supervised learning have been considered as AutoML at different times. Actually, any task trying to automate part of the machine learning design process can be considered AutoML. For instance, algorithm selection~\cite{RICE197665}, hyperparameter optimization~\cite{DBLP:conf/ijcnn/GuyonADB06}, meta-learning~\cite{10.1145/1456650.1456656,DBLP:journals/corr/abs-1810-03548}, full model selection~\cite{Escalante:2009:PSM:1577069.1577084}, 
Combined Algorithm Selection and Hyperparameter optimization (CASH)~\cite{ThoHutHooLey13-AutoWEKA}, neural architecture search~\cite{elsken2018neural}, etc. Because all of these tasks are closely related to each other, 
we refer to the unifying view proposed by Liu et al.~\cite{Zhengying-ads} instead. 

Liu et al. distinguish at least three levels of automation in which AutoML systems can be categorized, these are summarized as follows:\\
\begin{itemize}
    \item $\mathbf{\alpha-}$\textbf{level.} \emph{Search of estimators/predictors}.  This level refers to the task of defining / determining a function mapping inputs to outputs, for example, manually setting the weights of a linear regressor for approaching a particular task (here $y_i \in \mathbb{R}$), or generating hard-coded classifiers (e.g., based on if-then rules). 
    \item $\mathbf{\beta-}$\textbf{level.} \emph{Search of learning algorithms}. Refers to the task of determining the best learning algorithm for a given task.   Including methods that: 
    \begin{itemize}
        \item Explore the space of all estimators of a given class, e.g., hyperparameter optimization of a support vector machine (SVM) classifier~\cite{largemargin,learningwithkernels}. In this case, the form of $f$ is defined as: $f(\mathbf{x}) = sign( \sum_{j=1}^N \delta_j y_j k(\mathbf{x}_j,\mathbf{x})  + b )$, with $\delta$ denoting the variables associated to the Lagrange multipliers and $k$ an appropriate kernel function.  $\beta-$level methods in this setting could search for adequate kernel functions $k$ and additional hyperparameters for $f$ (e.g., regularizer term); likewise, these methods should still find the parameters of the model, e.g., $\delta$, $\mathbf{w}$ and $b$ values.
        \item Explore the space of all estimators that can be built from a set of learning algorithms and/or related processes like feature selection,  normalization if variables, etc. These type of $\beta-$level techniques include methods that automatically generate classification pipelines like:  PSMS~\cite{Escalante:2009:PSM:1577069.1577084} and Auto-WEKA~\cite{ThoHutHooLey13-AutoWEKA}. These methods are capable of determining the type of function $f$ (e.g., choosing an SVM or a decision tree classifier), but also they can specify additional procedures to be applied to the training data and/or the model, before, during or after $f$ is learned. For instance, typical processes could be: feature selection/extraction, building ensembles with partial solutions and adjusting the outputs of models (e.g., according to class imbalance rations). 
        $\beta-$level techniques are also in charge of determining the hyperparameters associated to any component of the \emph{full model}.

    \end{itemize}
    
        \item $\mathbf{\gamma-}$\textbf{level.} \emph{Search for meta-learning algorithms}. This level refers to methods that aim at exploiting a knowledge base of  tasks-solutions to learn to recommend/select $\beta-$level methods given a new task. This level includes techniques from the early meta-learning approaches for \emph{recommending} an algorithm from a number of options~\cite{DBLP:journals/air/VilaltaD02,10.1145/1456650.1456656}, to portfolio optimization methods~\cite{10.1145/2512962}, to surrogate models used in modern AutoML solutions~\cite{DBLP:journals/jmlr/GorissenDT09,Feurer2019},  
        to cutting edge few-shot meta-learning methodologies~\cite{DBLP:journals/corr/abs-1904-05046}.  Examples of $\gamma-$level AutoML techniques include   AutoSklearn that incorporates meta-learning as warm start for the optimization process~\cite{Feurer2019},  and  early AutoML solutions incorporating surrogates~\cite{DBLP:conf/cec/GorissenTCD08,DBLP:journals/jmlr/GorissenDT09}. The distinctive feature of $\gamma-$level approaches is that they take advantage of task-level information and use it for any aspect of the AutoML process. 
\end{itemize}

Under Liu et al.'s notion, most methodologies aiming to automate the design of machine learning systems can be covered~\cite{Zhengying-ads}. From the (manual) optimization of parameters for a fixed model, to the automation of any aspect of the design process. A remarkable feature of the above notion is that authors consider budgets (in time and space) for the different levels. Also, one should note that this notion is transverse to the categorization\footnote{Guyon et al. distinguish  methods  performing a search intensive procedure, called wrappers,  (mostly associated to $\beta-$level techniques),  those that are not data driven, called filters, (where $\gamma-$level methods can be framed) and embedded techniques (related to $\beta$ and $\alpha$ level methodologies).} of \emph{model selection} techniques  into filters, wrappers and embedded methods by Guyon et al.~\cite{10.5555/1756006.1756009, DBLP:books/sp/19/GuyonSBEELJRSSSTV19}.  
Please refer to~\cite{Zhengying-ads} for details and examples of tasks/methods falling under each of these categories. 

\subsection{Disentangling AutoML methods}
The field of AutoML has grown rapidly in the last few years and because of that a vast number of solutions are out there. In order to make it easier for the reader to distinguish across different AutoML techniques, in this section we describe the key components of any AutoML method. 

\begin{figure}[htb] 
	\centering    
	\includegraphics[scale=0.2]{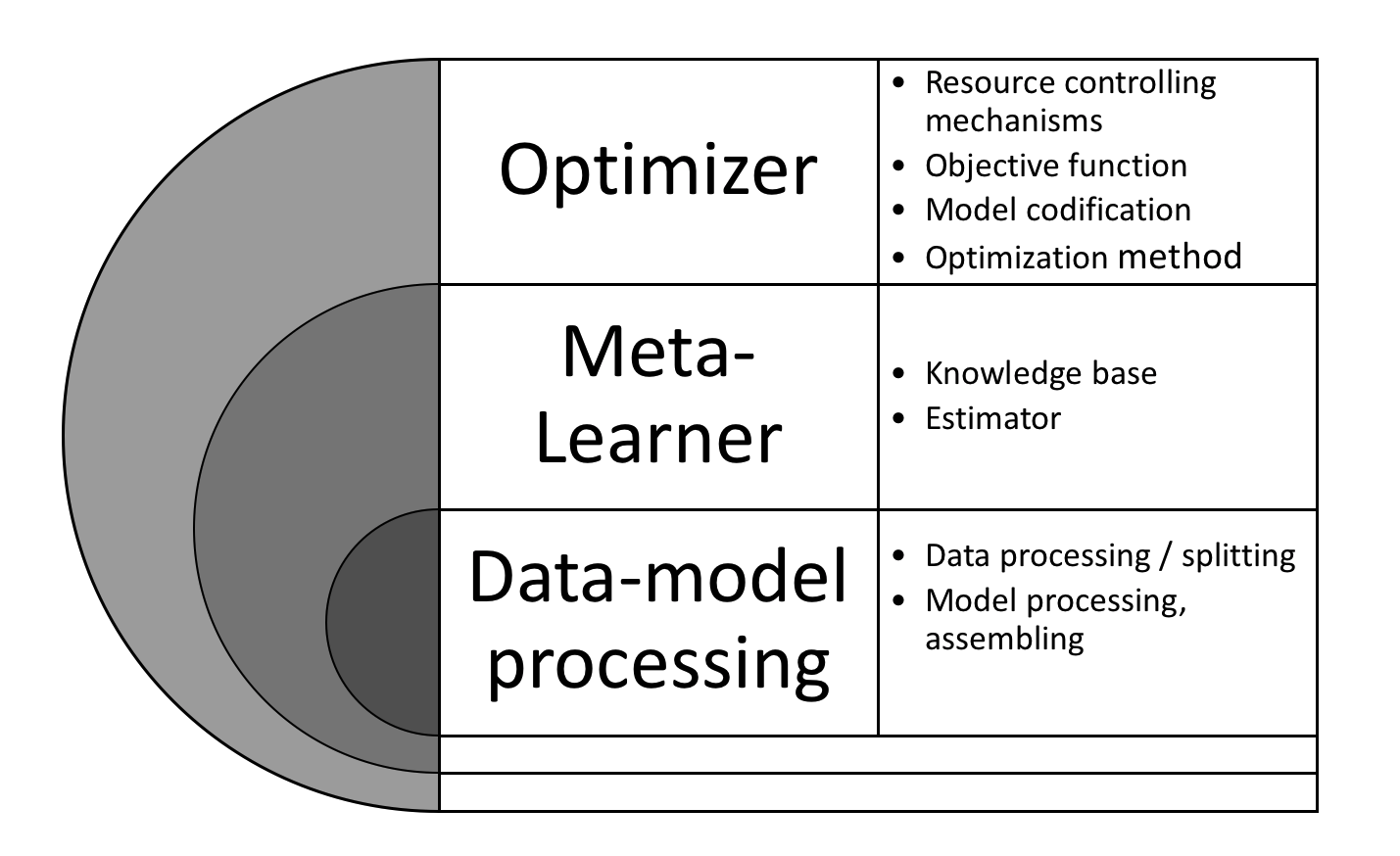} 
	\caption{Graphical diagram of the main components of an ($\gamma-$level) AutoML method. }\label{fig:disentanglement}
\end{figure}

In the author's opinion, one can distinguish three main components, namely:  \emph{Optimizer, Meta-learner} and \emph{data-model processing methods.} This categorization is graphically depicted in Figure~\ref{fig:disentanglement}.

The \textbf{optimizer} is the core of the AutoML method and it comprises the optimization algorithm itself, together with the objective function (usually a loss function for supervised learning). Resource controlling mechanisms are often associated to the optimizer, and the goal is to deal with the optimization problem while meeting time and memory budget constraints. 
Whereas generic optimization methods (e.g., evolutionary~\cite{DBLP:journals/jmlr/GorissenDT09,10.1145/2330784.2331014} and bio-inspired algorithms~\cite{Escalante:2009:PSM:1577069.1577084}, pattern search~\cite{doi:10.1137/1.9781611972726.16}, etc. ) have been traditionally used for this core component of AutoML, \emph{ad-hoc} optimization techniques tailored to the AutoML scenario are preferred. This include on a budget, anytime, and derivative free based methodologies. Likewise, multi objective techniques and methods that can operate over complex structures can have a positive impact in the overall performance of AutoML methodologies. Some of this methods are covered in other chapters of this book. 

A \textbf{meta-learner} refers to any estimator that is used during the AutoML optimization process, this could be a meta-learning technique for making recommendations on potentially useful models, or any other estimator (e.g., of expected performance or  running time)   used by the optimizer. Meta-learners are part of any $\gamma-$level approach. 
The meta-learner  is often coupled with the optimizer (e.g., in Auto-WEKA~\cite{ThoHutHooLey13-AutoWEKA} and AutoSklearn~\cite{Feurer2019})). 
\textcolor{black}{One should note that meta-learning by itself can be seen as an  AutoML methodology: early approaches were used to make \emph{coarse} model suggestions to solve supervised learning problems~\cite{DBLP:journals/air/VilaltaD02}. This form of algorithm recommendation/selection has been out there before the first AutoML formulations appeared. However, most of early meta-learning efforts focused on recommending a classification model, rarely  they also suggested hyperparameters. Therefore, full pipelines were not considered initially in meta-learning. Nowadays meta-learning is  a hot topic by itself, see~\cite{DBLP:journals/corr/abs-1810-03548,DBLP:journals/corr/abs-1904-05046}, and it has been \emph{synergistically} used in AutoML systems~\cite{Feurer2019}. We refer the reader to~\cite{DBLP:journals/corr/abs-1810-03548,DBLP:journals/corr/abs-1904-05046} for a complete review on meta-learning.  
}

\textbf{Data processing} mechanisms are those that modify, organize data according to the need of AutoML methods. These include data sampling and splitting for the assessment of solutions (e.g., successive halving methods~\cite{10.5555/3122009.3242042}). Finally, \textbf{model processing} techniques are those that enhance the model with \emph{ad hoc} mechanisms for improving AutoML solutions.  For instance, building ensembles with partial solutions like in~\cite{DBLP:journals/jmlr/GorissenDT09,DBLP:conf/ijcnn/EscalanteMS10,Feurer2019}. 

\textcolor{black}{Although this is not a strict categorization, most $\gamma-$level AutoML solutions adhere to it. Also, the interaction among these components is very flexible, for instance, there are AutoML methods that use the meta-learner before the optimization process while others use it during the search. This will be clearer in the next section where the most popular AutoML methodologies are described. }

\section{AutoML methodologies}
\label{sec:methods}
As previously mentioned, progress in AutoML has resulted in several methodologies that automate the design and development of supervised learning systems at different levels. It is out of the scope of this chapter to provide a complete review of existing methodologies, instead, in this section the most representative AutoML methodologies out there are described. The reader is referred to recent surveys in AutoML for complete description of the available methodologies~\cite{yao2018taking,zller2019survey,he2019automl,DBLP:books/sp/HKV2019,DBLP:journals/corr/abs-1907-08392,elsken2018neural}.
\begin{table}[htbp]
	\centering
	\caption{Overview of main AutoML methdologies shown in chronological order. Please note that $\alpha-$level methods are not included in this table as they refer to any methodology for fitting a model to a dataset (e.g., least-squares for linear regression). }
	\label{tab:historical}
	\footnotesize{
	\begin{tabular}{|p{1cm} |p{1cm} | p{1.7cm} | p{0.8cm} | p{3.5cm} |p{3.5cm} |}\hline
	\textbf{Year}& \textbf{Ref.}& \textbf{Method} & \textbf{Type} & \textbf{Description} & \textbf{Innovative aspects} \\\hline
	2006&\cite{psms06,Escalante:2009:PSM:1577069.1577084}& PSMS& $\beta$& Vectorial representation of solutions, PSO used as optimizer, subsampling, CV&Formulation of the full model selection task\\\hline
	2007&\cite{DBLP:conf/cec/GorissenTCD08,DBLP:journals/jmlr/GorissenDT09}&Heterogeneous surrogate evolution&$\beta$&Parallel co evolution of models, ensemble generation& Returned ensemble of solutions, large and heterogenous space of models\\\hline
	2010&\cite{DBLP:conf/ijcnn/EscalanteMS10}&Ensemble PSMS&$\beta$&Enhanced PSMS with ensemble of solutions& Returned an ensemble of solutions as output\\\hline
	2012&\cite{10.1145/2330784.2331014}&GPS: GA-PSO-FMS& $\beta$ & GAs were used to search for a model template, PSO was used for hyperparameter optimization&Separation of template search and hyperparameter optimization\\\hline
	2013&\cite{ThoHutHooLey13-AutoWEKA}&Auto-WEKA&$\gamma$&SMBO with SMAC, approached the CASH problem&Definition of the combined algorithm selection and hyperparameter optimization problem\\\hline
	2014&\cite{DBLP:journals/ijon/Rosales-PerezGCEG14}&Multi-objective surrogate-based FMS&$\gamma$&Multi objective (complexity/performance) evolutionary method, surrogates were used to approximate the fitness function&Among the first methods using a meta-learner for AutoML, multi-objective formulation\\\hline
	2015&\cite{Feurer2019}& AutoSkLearn & $\gamma$ & SMBO,  warm starting  with a classifier,  ensemble generation & AutoML definition, warm-starting with meta-learner, winner of AutoML challenge\\\hline
	2016&\cite{DBLP:conf/icml/OlsonM16,tpot}&TPOT&$\beta$&Genetic programming / NSGA-II selwection, cross validation, data sampling& Models naturally codified as GP trees \\\hline
	2017-2020&\cite{DBLP:journals/corr/ZophL16,autokeras,10.5555/3305890.3305981,elsken2018neural,Elsken2019}&Neural Architecture search &$\gamma$&Reinforcement Learning, Evolutionary Algorithms, SMBO for Neural Architecture search&Novel codifications for architectures, comparison of architectures, \emph{ad-hoc} NAS surrogates\\\hline
	\end{tabular}}
\end{table}

Table~\ref{tab:historical} summarizes the most representative methodologies of the early years of AutoML. These are shown in chronological order and the most important innovation or contribution from each methodology is briefly mentioned in the table. The goal of this table is just to provide a glimpse of the chronological development of AutoML. 
In the following we briefly describe some of these methodologies for further discussion below. We have divided this in waves that encompass methods that dealt with the problem similarly and that are chronologically  close to each other.

\subsection{First wave: 2006-2010}
Particle Swarm Model Selection (PSMS) is among the first existing AutoML methods dealing with the full pipeline generation problem~\cite{psms06,Escalante:2009:PSM:1577069.1577084}. Authors formulated the so called \emph{full model selection} problem, that consists of finding the best combination of data preprocessing, feature selection/extraction and classification models, together with the optimization of all of the associated hyperparameters. An heterogenous vector based representation was proposed to codify models into vectors and Particle Swarm Optimization (PSO) was used to solve the problem. A number of data sampling procedures were adopted to make the method tractable. In the same line, Gorissen et al. proposed a similar evolutionary algorithm to search for surrogates, where a wide variety of data preprocessing, feature selection/extraction and model postprocessing techniques could be considered to build the model~\cite{DBLP:conf/cec/GorissenTCD08}. To the best of the author's knowledge, this was the first work that proposed building ensembles as part of the AutoML  process. This notion inspired other methodologies like Ensemble PSMS~\cite{DBLP:conf/ijcnn/EscalanteMS10}, in which ensemble models of partial solutions found during the PSMS search process were returned as solutions. Building ensembles is nowadays part of most successful AutoML solutions like AutoSkLearn~\cite{Feurer2019}.  The last AutoML method from the early years of AutoML that we would like to mention is GPS~\cite{10.1145/2330784.2331014},  Quan et al. approached the full model selection problem with a quite novel formulation: in a first step, authors looked for a promising template for a classification pipeline and in a second stage authors optimized hyperparameters for the selected template. In the author's view this was a form of warm-starting the AutoML process, something that is common in contemporaneous AutoML solutions. 

From the above discussion, it is interesting that several core contributions widely used in current state of the art AutoML solutions were proposed during the first wave. Namely, a first formulation of the AutoML problem~\cite{Escalante:2009:PSM:1577069.1577084,DBLP:journals/jmlr/GorissenDT09}, the idea of building ensembles with information derived form the AutoML process~\cite{DBLP:journals/jmlr/GorissenDT09,DBLP:conf/ijcnn/EscalanteMS10} and initial ideas on warm-starting the search process~\cite{10.1145/2330784.2331014}. 

\subsection{Second wave: 2011-2016}
A second wave of AutoML started in the early 2010s with the introduction of models based on Bayesian Optimization / Sequential Model-Based Optimization (SMBO) for hyperparameter optimization and algorithm selection~\cite{snoek2012practical,smac}. The intuitive idea of these methods is to use a sort of surrogate  model to estimate the relations between performance and hyperparameters, and using this estimate to guide the optimization process via an acquisition function. In 2013 Thornton et al. introduced Auto-WEKA, an AutoML method based on SMBO capable of building classification pipelines in the popular WEKA platform~\cite{10.1145/1656274.1656278}. The authors formulated the CASH problem, which resembles similarities with the full model selection task. Auto-WEKA relied on a SMBO method called SMAC~\cite{smac} with a random forest estimator. 
This method boosted research in SMBO for AutoML that nowadays is the dominant optimization approach in this field. 

Interestingly,  alternative methodologies not adhering to a probabilistic formulation were proposed as well. For example, Rosales et al. developed an AutoML methodology based on multi-objective optimization and surrogate models~\cite{DBLP:journals/ijon/Rosales-PerezGCEG14}. A regressor (and later a classifier) was used to estimate the performance of solutions such that only the promising ones were evaluated with the costly objective function. This solution shares the spirit of SMBO but approaches the task in a  different way.

In the following years, solutions based on SMBO have been proposed, most notably AutoSklearn~\cite{Feurer2019}.  This method arose in the context of an academic challenge (see Section~\ref{sec:challenges}). AutoSklearn is based on SMBO with the distinctive feature that the search process is first initialized with a meta-learner that aims at directing the search towards promising models. Also, this method generates an ensemble of solutions explored during the search process. AutoSklearn won a series of AutoML challenges with large margin at some stages~\cite{DBLP:books/sp/19/GuyonSBEELJRSSSTV19}, and even outperformed humans that aimed to fine tune a model\footnote{During a live competition on manual model tuning that lasted a couple of days and was organized with WCCI2016~\cite{DBLP:books/sp/19/GuyonSBEELJRSSSTV19}. }. AutoSklearn was made publicly available and it is very popular nowadays.


\textcolor{black}{
Among the novel AutoML methodologies released after AutoSklearn is TPOT (Tree-based Pipeline Optimization Tool), a method based on evolutionary computation with a particular twist~\cite{tpot}. The distinctive feature of TPOT when compared to early efforts based on evolutionary computation is that TPOT uses genetic programming as optimizer, and models are coded as syntactic trees formed by primitives that correspond to models. Each tree represents a full classification pipeline and these are evolved to optimize performance while reducing the complexity (number of primitives used) of the pipeline. Codifying  pipelines as trees is a natural solution that has not been explored elsewhere.   }


\textcolor{black}{To summarize, the second waive can be credited by the emergence of Bayesian Optimization as the \emph{de facto} optimizer for AutoML, most AutoML solutions nowadays implement such modeling framework and differ in the way the estimators are defined or how they are used and coupled with other processes. This waive also witnessed the resurgence of meta-learning as a critical step towards automating the selection of classification models.  Likewise, progress in hyperparameter optimization resulted in techniques (e.g., multi fidelity approaches~\cite{DBLP:conf/aistats/JamiesonT16,10.5555/3122009.3242042,10.5555/2832581.2832731}) that have boosted research on  AutoML, see~\cite{Feurer2019HPO} for an up to date review on progress in this area.  }

\subsection{Third wave: 2017 and on}

\textcolor{black}{The current\footnote{One should note that efforts to automatically design neural networks arose in the early 90's, see, e.g.,~\cite{10.5555/93126.94034,10.1109/72.265960}. } trend in AutoML is that of techniques for Neural Architecture Search (NAS)~\cite{talbi:hal-02570804,Elsken2019,elsken2018neural,ren2020comprehensive,zller2019survey}. The outstanding achievements of deep learning across many fields, together with the enormous complexity that takes to manually tune a model to obtain the desired performance in a particular dataset has moved AutoML into the deep learning arena (in fact, several authors use as synonym AutoML with NAS). 
NAS deals with the problem of searching for the best architecture and hyperparameters of deep learning models.  Being this a very complex problem because of the number of associated parameters (of the order of billions) and the size of datasets that are required by these models to perform decently. 
Also, a  specific difficulty that must be addressed by NAS methodologies is 
the need for comparing heterogeneous structures (i.e., deep learning architectures). 
NAS is out of the scope of this chapter, however the reader is referred to~\cite{elsken2018neural,Elsken2019,talbi:hal-02570804,ren2020comprehensive} for  up to date surveys on this dynamic and fast evolving field. 
NAS together with few shot learning, and the use of reinforcement learning for AutoML processes comprise the third wave, great progress is expected in these fields as these  topics are in the  spotlight of the machine learning community. 
}



\textcolor{black}{This section has provided a broad review on the evolution of AutoML during the last decade. Although the review is not exhaustive, it gives the reader a clear idea on how the field has progressed and, most importantly, introduces the fundamentals of AutoML. In the remainder of this chapter we describe the role that challenges have had in the development of AutoML and we highlight open issues and research opportunities in the area.   }

\section{AutoML Challenges}
\label{sec:challenges}
It is well known that competitions have helped to advance the state of the art and to solve extremely complex problems that otherwise would have took much time, even centuries, see e.g.,  the \emph{Longitude Act}~\footnote{\url{https://worddisk.com/wiki/Longitude\_Act/}}. In the case of AutoML, they have played a major role, and, although it is arguable, in the authors' opinion, AutoML was born in the core of academic challenges. 
The  \emph{2006 Prediction challenge}~\cite{DBLP:conf/ijcnn/GuyonADB06} and the \emph{2007 agnostic learning vs. prior knowledge competition}~\cite{DBLP:conf/ijcnn/GuyonSDC07,DBLP:journals/nn/GuyonSDC08} challenged participants to develop methodologies that, with the less possible domain knowledge, could solve generic classification tasks.  This competition gave rise to a number of early AutoML solutions, see e.g.~\cite{DBLP:conf/ijcnn/Lutz06,DBLP:conf/ijcnn/CawleyT07,DBLP:conf/ijcnn/Reunanen07,DBLP:conf/ijcnn/Wichard07,DBLP:conf/ijcnn/EscalanteGS07}. Although most of them dealt with the hyperparameter optimization problem, for the first time  were assessed the advantages of including domain knowledge vs. developing completely agnostic methods when  building generic classifiers. The outcomes of the challenge brought light in that building an autonomous black box able to solve many classification problems was actually feasible. 
Please refer to~\cite{DBLP:journals/nn/GuyonSDC08,DBLP:conf/ijcnn/GuyonSDC07} for detailed analyses on the outcomes of the challenges and the developed solutions. 

The initial efforts of the previous competitions, were consolidated  years later throughout a series of challenges that were critical to boost the interest from the community in the AutoML field: \emph{the ChaLearn AutoML series}~\cite{DBLP:books/sp/19/GuyonSBEELJRSSSTV19,DBLP:conf/icml/GuyonCEEJLMRRSS16,DBLP:conf/ijcnn/GuyonBCEEHMRSSV15}. ChaLearn\footnote{\url{http://chalearn.org/}} lead the organization of a series of competitions that aimed at developing the \emph{dreamed AutoML blackbox} in a 5-stage evaluation protocol. Initially, participants dealt with binary classification problems, then  supervised learning problems of greater difficulty (regression, multiclass and multi-label classification)   were incorporated in subsequent stages. In each stage, five new datasets were released where participants did not know anything about the data, in fact, data was private remained  in the cloud until evaluation, so no code had access to the data beforehand. This was among the most novel feature of the AutoML challenge when compared to competitions at that time: solutions from participants of the AutoML challenge were evaluated autonomously in the cloud  without any user intervention.  Every AutoML solution was evaluated under the same conditions and using the same resources, it was during this challenge that budget restrictions were explicitly considered. The competition also allowed the comparison of pure AutoML solutions  with standard offline manually-tuned solutions. It was found that there was still a gap between fully autonomous vs. \emph{tweaked} approaches, motivating further research for the forth coming editions.  It is important to emphasize that it was in the context of this challenge that a popular and very effective AutoML method arose: AutoSklearn~\cite{Feurer2019}. 

The first edition of AutoML challenge series focused on mid size tabular data associated to supervised learning tasks. The complexity of the approached tasks was increased in the 
subsequent editions that focused on more realistic settings and more challenging scenarios. For instance, the Life Long AutoML challenge asked participants to develop solutions that could learn continuously in large scale datasets coming from real applications~\cite{10.1007/978-3-030-29135-8_8} and using non-standard data formats (e.g., temporal and relational data\footnote{\url{https://www.4paradigm.com/competition/kddcup2019}}). 
Because of the large scale of these datasets, solutions of the challenge focused on efficiency, hence, other aspects of AutoML were not targeted by participants (e.g., extensive search or overfitting avoidance mechanisms) 
Also,  the \emph{life long} setting motivated  participants to develop incremental solutions, in fact top ranked participants relied in boosting ensembles of trees, see e.g.,~\cite{10.1007/978-3-030-29135-8_13}. One of the most important outcomes of the  challenge was that efficiency in AutoML has not received enough attention from the community. Also, there was evidenced the lack of capabilities of state-of-the-art methods to handle non-tabular data. For a detailed description of the challenge please refer to~\cite{10.1007/978-3-030-29135-8_8}. 

The latest edition of the AutoML challenge series features the AutoDL\footnote{\url{https://autodl.chalearn.org/}} competition~\cite{liu:hal-02265053}. In this challenge,  participants are required to build AutoML methodologies able to work directly with raw data, where data can be heterogeneous (e.g., text, images, time series, videos, speech signals, etc.). Although the competition focus is on deep learning methodologies, any kind of method can be submitted. As in previous editions, solutions are evaluated in the CodaLab\footnote{\url{https://codalab.lri.fr/}} platform without any user intervention, methods do not have access to data until evaluation takes place, and there are budget restrictions. In preliminary evaluation phases of AutoDL that have focused in a single modality of data (e.g., images), very effective and efficient deep learning architectures have been already proposed~\cite{liu:hal-02386805}. Top ranked participants of these evaluation phases have developed efficient auto augmentation techniques~\cite{DBLP:journals/corr/abs-1905-00397} and have relied on \emph{light} architectures that are used as warmstart for the AutoML process (e.g., on mobile net~\cite{DBLP:journals/corr/HowardZCKWWAA17}). See~\cite{liu:hal-02265053,liu:hal-02386805,AutoDL2020} for details on the already existing methodologies that solve recognition tasks from raw data, without user intervention and by consuming reasonable resources. 

\textcolor{black}{This section has  provided an overview of academic competitions dealing with the AutoML problem under very different and challenging conditions.  By providing data, resources and accurate evaluation protocols, AutoML challenges have boosted research in different fronts of machine learning. From the \emph{2006 prediction challenge} to the {2020 AutoDL competition} the AutoML field has seen its rise within the machine learning community.  Several  effective methods for approaching the problem  have been proposed so far, some of them being widely used nowadays. AutoML is a clear example of what challenges can do, the field is growing with large portion of the machine learning community   actively working on it. 
}
\textcolor{black}{AutoML challenges have also made contribution to setting the basis for fair and standardized evaluations. Such evaluations are highly needed in AutoML, being a data driven and resource consuming process, ensuring autonomy, and delivering solutions within a reasonable time is critical. }

\section{Open issues and research opportunities}
\label{sec:openissues}
\textcolor{black}{In the last decade, AutoML has achieved a tremendous progress in trying to automate model design and development, mainly in the context of supervised learning. From initial efforts in trying to approach the problem with straightforward black-box optimizers based on vector representations, to the most recent studies aiming to compare graphs, and adopting meta-learning schemes.  With such a progress, the reader may be deceived that the AutoML task is solved (at least for tasks like classification), this, however, is still a far away goal. As there are several challenges that deserve attention from the community. In the following  some of the most promising problems for which research could make a tremendous impact are listed. }

\begin{itemize}
    \item \textbf{Explainable AutoML models.} AutoML solutions are, in general, black boxes that aim at exploring the space of models that can be build with a set of primitives. 
    Effective AutoML solutions are out there that can be used by any user without any formation in machine learning. Despite the progress, a direction that has not been explored by the community is that of developing \emph{transparent} AutoML methodologies. In the author's view AutoML models should be equipped with explainability and interpretability mechanisms.  This enhancement could bring important benefits for making AutoML accessible to everyone. Although this venue has not been explored yet, we are not far away from having transparent AutoML techniques, as  AutoML is in general a search intensive procedure that  generates vast amounts of information that can be exploited to generate explainable and interpretable AutoML solutions.

    \item \textbf{AutoML in feature engineering.} Although data processing, including feature selection and extraction, have been considered as components of pipelines generated by AutoML techniques, the feature engineering process by itself has received little attention from the community. It is only recently that efforts aiming to process raw data directly are emerging, see~\cite{UKhurana,10.5555/3172077.3172240,liu:hal-02265053,liu:hal-02386805,DBLP:conf/ciarp/MadridE19}. We believe this research venue will be decisive for the full automation of the AutoML process. 
    
    \item \textbf{AutoML for non tabular data.} Related to the previous point, AutoML methods for dealing with non tabular data, including raw data (e.g., text, images, etc.) and structured data (e.g., graphs, networks, etc.) are becoming more and more necessary, hence this area could also be a fruitful venue for research. 
    
    \item \textbf{Large scale AutoML.} \textcolor{black}{Large scale problems are still an open problem for state of the art AutoML solutions. This was evidenced in recent AutoML challenges were only few solutions could perform search intensive AutoML procedures~\cite{10.1007/978-3-030-29135-8_8}. This represents  an open problem that deserves further attention from the community. Likewise, in  deep learning models, AutoML has to be efficient and there are already several \emph{efficient} implementations of NAS models.}

    \item \textbf{Transfer learning in AutoML.} As previously mentioned, current AutoML solutions generate vast and rich information that can be useful for a number of purposes. A promising purpose for taking advantage of such information is to perform transfer learning to enhance the performance of AutoML models. Meta-learning procedures already perform a sort of transfer learning, however, a promising research venue is to transfer knowledge on information on the optimization process (e.g., transferring information on the dynamics of the optimization process from task to task).
    
    \item \textbf{Benchmarking and reproducibility  in AutoML.} Since AutoML is an optimization process that involves data, efforts on developing platforms and  frameworks for the evaluation and fair comparison among AutoML methodologies is an open issue that is gaining attention from the ML community (see, e.g.,~\cite{lindauer2019best}). Whereas challenges offer such platform, they may become obsolete rapidly given the speed at which AutoML is growing. Likewise, code sharing and mechanisms that encourage the reproducibility of results in AutoML could have a huge positive impact in the matury of the field.
    
    
    \item \textbf{Interactive AutoML methods. } While the main goal of AutoML is to \emph{automate} processes and removing as much as possible to the user from design loop, interactive AutoML methodologies could take the performance of AutoML models far away from its current status. Mechanisms for including prior knowledge into the AutoML process could have a positive impact. 
    
\end{itemize}

\section{Conclusions}
\label{sec:conclusions}
Automated machine learning aims at helping users with the design of machine learning systems.  From the optimization of hyperparameters of fixed models, to model type selection and full model/pipeline generation to the automatic design of deep learning architectures, AutoML is now an established field with wide applicability in the \emph{data science era}.  
Great progress has been achieved in the early years, with very effective methodologies readily to use for users with limited knowledge in machine learning. Likewise, solutions making 
\emph{easier} the design task even for machine learning experts. 

This chapter has provided an overview of the 
major achievements during this first decade, the most representative methodologies were presented and the fundamentals of the AutoML task were provided.  Perhaps the most important conclusion one can draw from this early years of progress is that nowadays we have evidence that AutoML is a feasible task, this is a very important result as in the early years the machine learning community was very skeptical on the future of the field. Hence, even when the black-box all-problem solution is far from being reached, today we can leverage on AutoML techniques to approach problems that traditionally required of considerable effort. Also, it is clear the role that AutoML challenges have had into the establishment of the field. 
With the progress seen in this first decade, much is expected from AutoML in the next few years. In particular, it will be exciting to know of methodologies that can approach the open problems highlighted in the previous section. Also, it is intriguing to what extend will it be possible to take automation in deep learning.  

\section*{Acknowledgments}
The author is grateful with the editors for their support in the preparation of this manuscript. The ChaLearn collaboration (\url{http://www.chalearn.org/}) is acknowledged for the organization of the ChaLearn AutoML challenges series. This research was partially supported by CONACyT under project CB-26314: \emph{Integraci\'on de Visi\'on y Lenguaje mediante Representaciones Multimodales Aprendidas para Clasificaci\'on y Recuperaci\'on de Im\'agenes y Videos.}
\bibliographystyle{plain}
\bibliography{paper}
\end{document}